\documentclass[10pt,twocolumn,letterpaper]{article}

\usepackage{iccv}
\usepackage{times}
\usepackage{epsfig}
\usepackage{graphicx}
\usepackage{amsmath}
\usepackage{amssymb}

\usepackage{graphicx}
\usepackage{amsmath}
\usepackage{amssymb}
\usepackage{booktabs}
\usepackage{verbatim}
\usepackage{pifont}%
\usepackage{multirow}
\usepackage{subcaption}
\usepackage{acronym}
\usepackage{xcolor}
\usepackage{color}

\usepackage[numbers,sort]{natbib}

\usepackage[pagebackref=true,breaklinks=true,letterpaper=true,colorlinks,bookmarks=false]{hyperref}

\iccvfinalcopy %

\newcommand{\fig}[1]{Figure~\ref{#1}}

\newcommand{\tbl}[1]{Table~\ref{#1}}

\def\eg{\emph{e.g.}} 
\def\ie{\emph{i.e.}}

\def\etal{\emph{et al.}}

 \newcommand{\supp}{supplementary materials\xspace}

\newcommand{\R}{\mathbb{R}}

\newcommand{\z}{\mathbf{z}}

\newcommand{\T}{T}
\newcommand{\K}{S}
\newcommand{\Lv}{N}
\newcommand{\nar}{n}
\newcommand{\vis}{v}
\newcommand{\wkh}{s}
\acrodef{MSTAN}[\textsc{VINA}]{Video, Instructions, and Narrations Aligner}
\newcommand{\myparagraph}[1]{\smallskip\noindent\textbf{#1.}}

\newcommand*\samethanks[1][\value{footnote}]{\footnotemark[#1]}

\ificcvfinal\fi

\begin{document}

\title{Learning to Ground Instructional Articles in Videos through Narrations}

\author{Effrosyni Mavroudi\thanks{equal contribution}, \ 
Triantafyllos Afouras\samethanks, \
Lorenzo Torresani \\
Meta AI \\
{\tt\small \{emavroudi, afourast, torresani\}@meta.com}
}

\maketitle
\ificcvfinal\thispagestyle{empty}\fi

\begin{abstract}
In this paper we present an approach for localizing steps of procedural activities in narrated how-to videos. To deal with the scarcity of labeled data at scale, we source the step descriptions from a language knowledge base (wikiHow) containing instructional articles for a large variety of procedural tasks. Without any form of manual supervision, our model learns to temporally ground the steps of procedural articles in how-to videos by matching three modalities: frames, narrations, and step descriptions. Specifically, our method 
aligns steps to video by fusing information from two distinct pathways: i) {\em direct} alignment of step descriptions to frames, ii) {\em indirect} alignment obtained by composing steps-to-narrations with narrations-to-video correspondences. 
Notably, our approach performs global temporal grounding of all steps in an article at once by exploiting order information, and is trained with step pseudo-labels which are iteratively refined and aggressively filtered.
In order to validate our model we introduce a new evaluation benchmark -- HT-Step -- obtained by manually annotating a 124-hour subset of HowTo100M\footnote{A test server is accessible at \url{https://eval.ai/web/challenges/challenge-page/2082}.} with steps sourced from wikiHow articles. Experiments on this benchmark as well as zero-shot evaluations on CrossTask demonstrate that our multi-modality alignment yields dramatic gains over several baselines and prior works. Finally, we show that our inner module for matching narration-to-video outperforms by a large margin the state of the art on the HTM-Align narration-video alignment benchmark. 
\end{abstract}

\section{Introduction}

\begin{figure}[t]
    \centering
    \includegraphics[width=\columnwidth, trim={1cm 1cm 1cm 0},clip]{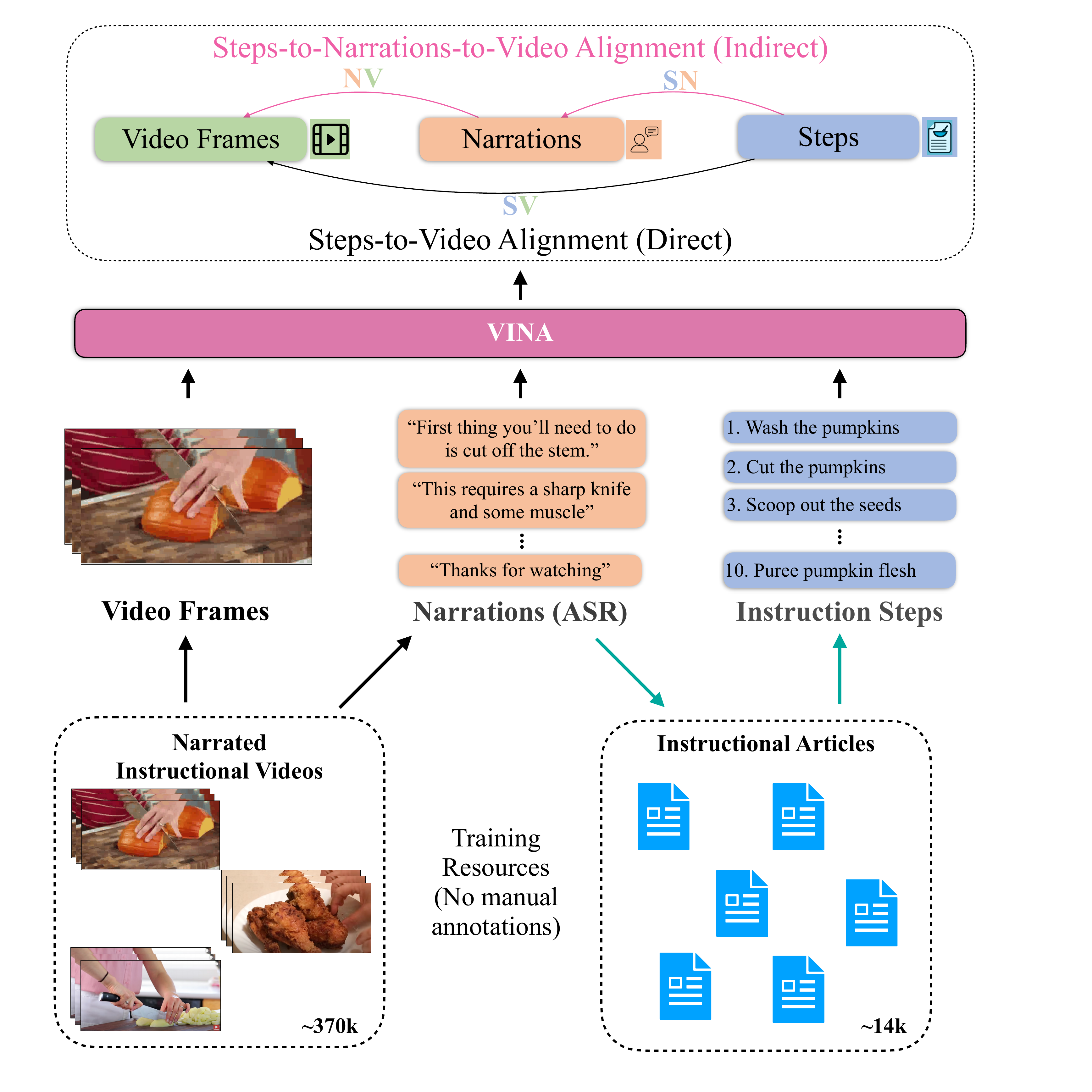}
    \caption{ 
     Our proposed \ac{MSTAN} learns to simultaneously ground narrations and instruction steps in how-to videos from an uncurated set of narrated videos and a separate knowledge base of instructional articles, \emph{without any manual annotations}. This is contrary to prior work that learns how to align a video with a \emph{single} sequence of sentences by leveraging \emph{ground-truth pairs} of video-text sequences, e.g., a video and its narrations~\cite{Han_CVPR22}, or a video and an annotated, ordered list of steps demonstrated in it~\cite{Dvornik2021DropDTWAC}.
    }
    \vspace{-0.5cm}
    \label{fig:teaser}
\end{figure}

Instructional videos have emerged as a popular means for people to learn new skills and improve their abilities in executing complex procedural activities, such as cooking a recipe, performing home improvements, or fixing things. In addition to being useful teaching materials for humans, how-to videos are a promising medium for learning by machines, as they provide revealing visual demonstrations of complex activities and show elaborate human-object interactions in a variety of domains. Motivated by this observation, in this work we look at the task of temporally localizing the steps of procedural activities in instructional videos. This problem is foundational to the broader goal of human-procedure understanding and advances on this task promise to enable breakthrough applications, such as AI-powered skill coaching and human-to-robot imitation learning. 

Prior work has tackled procedural step localization by leveraging either (a) fully-annotated datasets where the task shown in the video is given (\emph{video-level labeling}) and manually annotated temporal segments are provided for each step (\emph{segment-level labeling})~\cite{tang19coin} or (b) weakly-annotated training sets where the task and the order in which the steps appear in the video is given~\cite{Zhukov_CVPR19}. However, due to the inherent manual cost involved in collecting step annotations, these works have relied on datasets that are small-scale both in the number of tasks (e.g., at most few hundreds~\cite{Zhukov_CVPR19}) and in the number of video samples (e.g., 12k videos~\cite{tang19coin}). These limitations affect both the generality and the complexity of the models that can be trained on these benchmarks. In this paper, we therefore pose the following question: \emph{can we leverage large-scale, unlabeled video datasets to train a model that can ground procedural steps in how-to videos}? 

To answer this question, we propose a novel training framework for weakly-supervised step grounding that utilizes two freely available sources of information: (a) instructional articles which define ordered lists of steps for a wide variety of tasks (e.g., from wikiHow) and (b) narrations which provide instance-specific rich commentaries of the execution of the task in the video, e.g., from ASR transcriptions. Our work treats the former as an abstraction of the latter and uses the video-specific narrations to support the grounding of the article steps. Specifically, {\em during training}, our method leverages narrations as an auxiliary signal to (i) identify the task shown in the video, (ii) temporally ground the article steps that are visually-demonstrated and (iii) filter out steps that are not executed in the given instance. To further motivate this mechanism, let us look at the example in \fig{fig:teaser}. The narrations help disambiguate the task (\emph{make a pumpkin puree}), enabling the automatic retrieval of relevant instructional articles for the video. Furthermore, the narrations can be matched to steps described in the articles to roughly localize the steps that are represented in the video. In this example, the timestamp of ``First thing you’ll need to do is cut off the stem” provides a loose temporal prior for the matching step ``Cut the pumpkins.''  On the other hand, steps that do not have any matching narrations (e.g., ``Wash the pumpkins'') are unlikely to be represented in the video and thus can be rejected. Based on this intuition, we propose a procedure that learns to align steps to video by fusing information from two pathways. The first is an {\em indirect} pathway inferring step-frame alignments by 
composing step-to-narration assignments with narration-to-frame correspondences. The second is a {\em direct} pathway that learns associations between step descriptions and frames by leveraging information from all videos having steps in common. 

In our experiments we demonstrate that our multi-modality alignment leads to significant performance gains over several baselines, including single-pathway temporal grounding, as well adaptations of prior works to our problem. {\em During inference}, the direct pathway can be used by itself to temporally ground steps in absence of transcribed narrations. When narrations are available at test time, our method improves further the accuracy of temporal grounding by fusing the inference outputs of the two pathways. 

To summarize, our work makes the following contributions:
1) we learn to align steps to frames in how-to videos, using only weak supervision in the form of noisy ASR narrations and instructional articles;
2) we propose a novel approach for joint dense temporal grounding of instructional steps and video narrations;
3) we introduce a new benchmark for evaluating instructional step grounding
which we will make available to the community;
4) we demonstrate state-of the art results on multiple benchmarks for both step as well as narration grounding. 
\section{Related Work}
\paragraph{Procedural step recognition.}

Prior work on procedural step localization~\cite{rouditchenko2020avlnet,CaoOtam,shvetsova2022everything,ko2022video,cao2022locvtp,Elhamifar20,p3iv22,bi21,pdpp23,Zhong23,yang2021temporal,shen2021learning,Zhukov_CVPR19,Luo2020UniVL} can be roughly divided
into two categories, based on the query formulation:
the first class approaches the problem in an open-world setting,
where the use of text queries transforms it into a temporal grounding task~\cite{anne2017localizing,bao2021dense,ging2020coot}.
Such approaches can be further sub-divided into single step grounding, where single steps are queried over the whole video\cite{Sun2019VideoBERTAJ,Kuehne16end} and dense temporal grounding methods~\cite{Han_CVPR22,Chen_EMNLP22} where the objective is to jointly ground a sequence of steps or whole article into the video. 
The second body of works uses fixed taxonomies of steps, often as part of activities~\cite{tang19coin,shen2022semi,lea2016temporal}. 
Our work is somehow related to Lin et al.~\cite{lin2022learning} who use semantic similarity between 
steps and narrations to obtain supervision for learning strong video representations. Although we also associate steps from wikiHow articles to video frames through the use of narrations, the two works differ in several aspects: we align steps to video by a global procedure that takes into account all ordered steps in the article (inspired by dense temporal grounding methods~\cite{bao2021dense}) and temporally grounds them in the whole video, instead of matching individual video clips to an orderless collection of steps; our step grounding uses {\em video} in addition to steps and narrations while the method proposed in Lin et al. relies purely on text-matching narrations to step descriptions; finally, the works differ in objectives with our aim being step grounding in long how-to videos rather than learning video-clip representations.

Existing methods also vary by the level of supervision used during training.
One option is leveraging  fully-annotated datasets with known temporal segments for each step~\cite{ziani2022tempclr,cao2022locvtp,shen2022semi,lea2016temporal,kuehne2014language,singh2016multi,tang2020comprehensive,chinayi_ASformer},
using weakly-annotated training sets where the task and the order in which the steps appear in the video are known~\cite{Zhukov_CVPR19,Fried:ACL20,Dvornik2021DropDTWAC,Chang2019D3TWDD,bojanowski2014weakly,bojanowski2015weakly,richard2018neuralnetwork,kuehne2017weakly}, only the task and potential steps are known~\cite{richard2018action},
or only loose association  between video and instructional articles is given~\cite{Chen_EMNLP22,Dvornik:ECCV22}. Video narrations are a commonly used source of weak supervision~\cite{alayrac2016unsupervised,malmaud2015whats,sener2015unsupervised,Fried:ACL20}, while instructional steps from knowledge bases have recently been used as supervision:~\cite{lin2022learning,Chen_EMNLP22}.
Chen et al. ~\cite{Chen_EMNLP22} use video-level instructional step labels for (weak) supervision of a model that grounds instructional articles to videos. This approach attempts to localize steps without using any narration information; we instead show that the task knowledge is not necessary and heavily exploit narrations via multi-task learning and complementary inference pathways~\cite{zamir2020robust}: we argue that narrations provide a much richer source of supervision for training step grounding models, while essentially coming for free.
\vspace{-0.5cm}
\paragraph{Video-Text alignment}
The availability of large-scale video-text datasets such as HowTo100M has prompted
many works on joint video-language embedding training~\cite{miech20endtoend,lin2022learning}.
A form of contrastive loss is often adopted for bringing together the representations of the two modalities~\cite{miech20endtoend,xu-etal-2021-videoclip,ma2022simvtp,Ma2022XCLIP,park-etal-2022-normalized,Yang2021TACoTC,bain2021frozen,radford2021learning}, while
 masked objectives are also gaining popularity~\cite{Sun2019VideoBERTAJ,Zhu20actbert,ma2022simvtp,tong2022videomae,tan2019lxmert,huang2019unicoder,chen2020uniter,Su2020VL-BERT,li2020hero}.
Some works perform end-to-end representation learning~\cite{miech2019howto100m,miech20endtoend}, while others freeze representation and focus on longer-term temporal modelling, which aims to capture context~\cite{xu-etal-2021-videoclip}. 
More recently Han~\etal~investigated directly aligning contextualized narration representations to video frames~\cite{Han_CVPR22}. 
We build our method off of this approach -- we note however that our objective is complementary: rather than aligning a video's narrations as an end-goal, we use this functionality to ground a set of independent steps sourced from instructional articles; in that process we show that the synergy that develops while training jointly on the two tasks results in improved performance for both. 
\section{Narration-Aided Step Grounding}
We first present our architecture for joint narration and step grounding (Sec.~\ref{subsec:method_arch}), followed by learning objectives (Sec.~\ref{subsec:objectives}) and pseudo-labeling strategy (Sec.~\ref{subsec:method_pseudolabeling}); we discuss inferring the video task in (Section ~\ref{subsec:method_task_inf}).

\subsection{Problem Formulation}
\label{subsec:method_prob}
Let ($\mathcal{V}, \mathcal{N}$) be a video-narration pair, consisting of $\T$ video frames and a sequence of $\Lv$
narrations. 
Also, let $\mathcal{S}$ be an ordered list of $S$ steps from an instructional article for a candidate task $\tau$. Our objective is to ground each step of $\mathcal{S}$ to the video, conditioned on the other steps and the ASR transcript\footnote{ASR transcripts are assumed to be always available for training and optionally during inference.}.
In particular, the desired output of our model is an alignment matrix $Y^{SV} \in \{0,1\}^{S \times T}$, where $Y_{st}=1$ only if frame $t$ is depicting the $s$-th step of task $\tau$, and zero otherwise. Note that some steps might not be represented in the video.

\subsection{Joint Narration and Article Step Grounder}
\label{subsec:method_arch}
\begin{figure*}[ht]
    \centering
    \includegraphics[scale=0.25, trim={0cm 2.5cm 0cm 1cm},clip]{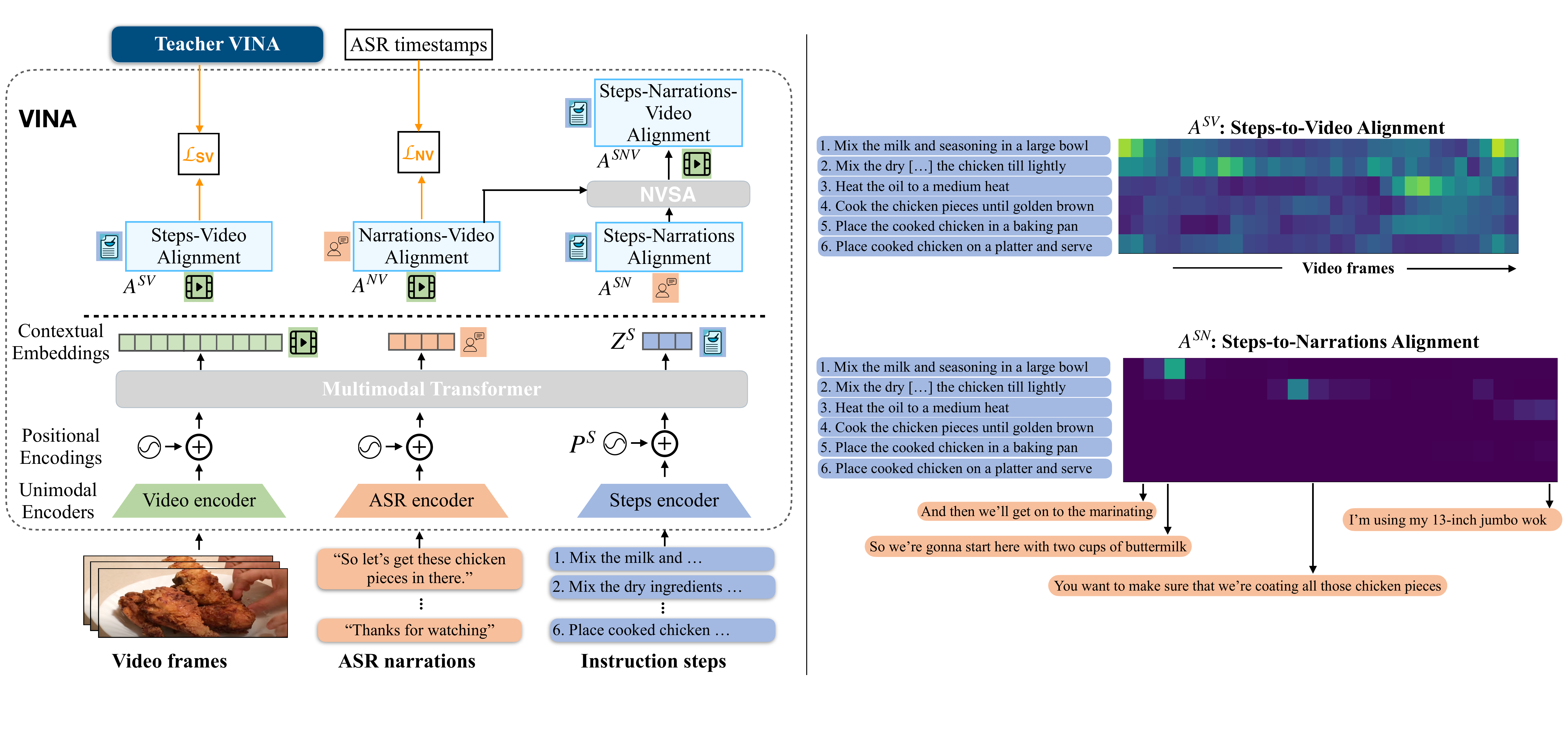}
    \caption{(\emph{left}) Schematic illustration of our system. First, it extracts representations for each modality (video, ASR narrations, task steps) with three \emph{Unimodal Encoders}. These representations are fed to a Transformer-based \emph{Multimodal Encoder} for capturing interactions among video frames as well as among the textual modalities and the video. 
    The contextualized embeddings for video frames, narrations 
    and steps are used to compute correspondences between \emph{all pairs of modalities}.  
    Step grounding is achieved by fusing the output of two pathways: a {\em direct} pathway aligning steps to video ($A^{SV}$) and an {\em indirect} pathway that composes steps-to-narration ($A^{SN}$) with narration-to-video ($A^{NV}$) alignments to produce a second step-to-video alignment ($A^{SNV}$). We train our model using a teacher-student strategy with iteratively refined and filtered step pseudo-labels. (\emph{right}) Qualitative examples of our learned steps-to-video alignment and steps-to-narrations alignment matrices for a video snippet.}
    \label{fig:method}
\end{figure*}
As shown in Figure~\ref{fig:method}, our proposed \ac{MSTAN} model follows the 
popular paradigm of leveraging Transformers for modeling multimodal 
interactions~\cite{tsai-etal-2019-multimodal,Peng22}.

\myparagraph{Unimodal Encoders} Before feeding the video, narrations and 
article steps to our model, we preprocess them to extract a sequence of 
tokens.
Given a video-narration pair ($\mathcal{V}$, $\mathcal{N}$) 
we extract visual features,
$X^{\vis} \in \R^{T \times D_{\vis}}$ 
and narration features
 $X^{\nar} \in \R^{N \times D_{\nar}}$ using standard backbone networks (e.g., a frozen S3D~\cite{xie2018rethinking} network for visual features, and pooled Word2Vec~\cite{mikolov13} embeddings for narration features).
 Similarly, we encode the sequence of 
steps in a sequence of features $X^{\wkh} \in \R^{S \times D_{\wkh}}$. 
The features of each modality $m$ are embedded into a 
common embedding space of dimensionality $D$ using a Unimodal Encoder that consists of a modality-specific MLP network, and then 
learnable, modality-specific positional embeddings
$P^{m}$ are added to them:
\begin{align}
H^{m} = MLP(X^{m};\theta_{m}) + P^{m},
\end{align}
where $m \in \{\vis,\nar,\wkh\}$ denotes the modality.

\myparagraph{Multimodal Encoder} The outputs of the Unimodal Encoders are 
concatenated into a sequence of tokens:
$H = [H^{\vis}; H^{\nar}; H^{\wkh}] \in \R^{(\T + \Lv + \K) \times D}$
and fed to the Multimodal Encoder, which is a 
standard Transformer with multiple layers of multi-head self-attention:
\begin{equation}
Z = \mathrm{Transformer}(H) \in \R^{(\T + \Lv + \K) \times D}.
\end{equation}
The contextualized embeddings  $Z=[Z^{\vis}; Z^{\nar}; Z^{\wkh}]$ computed by the 
Multimodal Encoder capture interactions within each modality (e.g., temporal 
relationships within the video and context among steps of an  article) and 
across modalities.
We can then compute cosine similarity matrices between all pairs of modalities: narrations-to-video $A^{NV} \in \R^{N \times T}$, steps-to-video $A^{SV} \in \R^{S \times T}$, and steps-to-narrations $A^{SN} \in \R^{S \times N}$. For example, the narrations-to-video similarity matrix $A^{NV}$ is obtained by simply computing the cosine similarity between each frame embedding and each narration embedding:
$
A^{NV}_{nt} = {\z^{\nar}_n}^\top \z^{\vis}_t / \left( ||\z^{\nar}_n|| \ ||\z^{\vis}_t|| \right)
$.

\noindent \myparagraph{Narration-aided Step Grounding} A straightforward inference path for temporally grounding the steps in the video is directly through the $A^{SV}$ similarity matrix, which captures the similarity of each video frame with each instructional step.
However, this alignment does not explicitly take into account the narrations of the video (only implicitly, through the Multimodal Transformer).
We observe that an alternative way to ground steps in a video is to first identify narrations in the ASR transcript that are relevant to the step and then exploit the similarity of those narrations with video frames to get a loose prior over the step location.
This is computed by combining the information captured in the steps-to-narrations and narrations-to-video alignment matrices $A^{SN}$ and $A^{NV}$:
\begin{equation}
    A^{SNV} = \tilde{A}^{SN} A^{NV} \in \R^{S \times T},
\end{equation}
where $\tilde{A}^{SN}$ is the predicted steps-to-narrations alignment matrix $A^{SN}$ after being normalized with a softmax function with temperature $\xi$:
$\tilde{A}^{SN}_{sn} = \frac{\exp{(A_{sn} / \xi)}}{\sum_{j=1}^N \exp(A_{sj} / \xi)}$.

The resulting $A^{SV}$ and $A^{SNV}$ alignment matrices provide two complementary inference paths to align steps to video frames. The mutual agreement between the direct steps-to-video 
alignment provided by $A^{SV}$ and indirect, narration-based steps-to-video alignment provided by $A^{SNV}$ can be used to better ground steps. Intuitively, if a frame is both very similar to a step in the joint embedding space learned by the Multimodal Transformer, and also very similar to a narration that is relevant to the step, then it is more likely to be indeed relevant to the step.
Hence, we fuse the $A^{SV}$ and $A^{SNV}$ alignment matrices to a matrix $A^{F} = (A^{SV} + A^{SNV})/2$.

\subsection{Weakly-Supervised Training from Narrated Instructional Videos} 
\label{subsec:method_training}
Next, we discuss how to supervise the \ac{MSTAN} model in order to learn steps-to-video alignment and narrations-to-video alignment. We first present the training objective assuming that the ground-truth temporal segments for each narration and step in the video are given.
Then we describe our approach for obtaining automatic pseudo-labels for the temporal segments.
\vspace{-0.3cm}
\subsubsection{Learning on Labeled Data}
\label{subsec:objectives}
Let ${\mathcal{B} = \{\mathcal{V}_i,\mathcal{N}_i,\mathcal{S}_i,Y^{NV}_i, Y^{SV}_i\}_{i=1}^{B}}$ denote a set of training tuples,
each comprising a 
video-narration pair, an ordered list of relevant task steps, 
and the target video-narrations and video-steps alignment matrices,
we train the \ac{MSTAN} model by optimising the following objective:
\begin{align}
\mathcal{L} & =  \frac{1}{B} \Bigl[ \sum_{i=1}^B \lambda_{NV} \mathcal{H}(Y_i^{NV}, A_i^{NV}) + \lambda_{SV} \mathcal{H}(Y_i^{SV}, A_i^{SV})  \Bigr], 
\end{align}
where $\mathcal{H}(\cdot,\cdot)$ is the modified InfoNCE loss
used by~\cite{Han_CVPR22} for aligning video with 
text using noisy ground-truth temporal segments:
\begin{align}
\mathcal{H}(Y,A) = - \frac{1}{K} \sum_{k=1}^K \log \frac{\sum_{t} Y_{t,k} \exp{(A_{t,k}/ \eta)}}{\sum_{t} \exp{(A_{t,k}/ \eta)}},
\end{align}
where
$\eta$ is a temperature constant.
Note that although we do not explicitly supervise the steps-narrations alignment $A^{SN}$,
meaningful alignments emerge during training due to the joint grounding of narrations and steps to the same video samples, as seen in \fig{fig:method}. Note that although we do not directly supervise the steps-to-narrations alignment, our model is able to learn meaningful correspondences, which go beyond simple pairwise textual matching.

\subsubsection{Pairing Videos with Articles}
\label{subsec:method_task_inf}
We assume access to a set of instructional articles  $ \mathcal{A} = \{\mathcal{S}_j, \tau_j\}_{j=1}^W$,
where $\tau_j$ denotes the article title and $\mathcal{S}_j$ the associated set of steps. 
To assign a set of steps to a given video from our training set $\mathcal{B}$ we need to associate it with an article from $\mathcal{A}$. 
If our video dataset provides metadata (e.g., a task id for every video), then this can be used to obtain the association -- although there is no guarantee that this will result in the best article-match for the video (see discussion in \supp for more analysis). 
If such metadata is not available, we can predict a task id, using the similarity between the narration and the titles of the available articles.
To that end we use an off-the-shelf language model (e.g. MPNet~\cite{song2020mpnet})
to compute semantic embeddings of the ASR captions of every video and the title of each article $\tau_j \in \mathcal{W}$. 
For every video $\mathcal{V}_i$ we then calculate the semantic similarity between all the $N$ captions in $\mathcal{N}_i$ and all task titles $\tau_j \in  \mathcal{W}$,  
and assign $N$ votes; the vote of every caption goes to the task that best matches it. 
Finally the video is assigned the task with the most votes.
Alternatively, in order to obtain multiple sets of steps for a video, we rank the tasks by the number of votes. 

\subsubsection{Narration-Aided Pseudo-Labeling}
\label{subsec:method_pseudolabeling}
Once a task $\tau_j$ has been associated with a video, we have access to a list of steps $\mathcal{S}_j$ from the article of the task. 
However, whether these steps appear in the video and their temporal location remain unknown. Inspired by self-labeling approaches from the SSL literature~\cite{lee2013pseudo,sohn2020fixmatch,wang2022semi}, we follow a teacher-student approach where a teacher version of our models generates pseudo-labeled temporal segments for training the student. 
For every step represented by a row in the learned steps-to-video alignment matrix we obtain a pseudo-ground truth
segment by finding the maximal activation (peak) and expanding
a temporal segment on both sides until the activation falls below an adaptive threshold $\zeta$ (e.g., 70\% of the peak).
To avoid training with unreliable pseudo-labels, we filter out pseudo-labels with low confidence: if the peak activation is below a fixed threshold $\gamma$, the alignment of that step is treated as unreliable for pseudo-labeling, and is altogether ignored.

\myparagraph{Training curriculum} For the first $E_b$ epochs we perform burn-in training of the student model on fixed pseudo-labels generated by feeding the video and the list of steps $\mathcal{S}_j$ to TAN~\cite{Han_CVPR22}, an off-the-shelf model pre-trained on the task of video-text alignment. 
Afterwards, we switch to using pseudo-labels generated from the teacher, where the teacher is initialized by duplicating the burn-in student model and then updated every $\nu$ epochs. During both stages, we utilize the original ASR timestamps for supervising the video-to-narrations alignment.
\section{Experiments}
\subsection{Datasets and Metrics}
We train our models on narrated videos from the HowTo100M dataset by leveraging the dataset release of wikiHow instructional articles~\cite{Koupaee2018wikiHowAL}, without using any form of manual annotations. In order to evaluate the effectiveness of our method, we evaluate: step grounding on HT-Step (a new benchmark, described below), narration alignment on HTM-Align~\cite{Han_CVPR22}, and zero-shot step localization on CrossTask~\cite{Zhukov_CVPR19}. 

\myparagraph{HowTo100M (Training)} 
The HowTo100M dataset~\cite{miech2019howto100m} contains instructional videos from YouTube. Following Han~\etal~\cite{Han_CVPR22}, we use the Food \& Entertainment subset containing approximately $370K$ videos, where each video is complemented by the ``sentencified'' ASR transcription of its audio narration.

\myparagraph{wikiHow (Training)} We train using 14,541 cooking tasks from the wikiHow-Dataset~\cite{Koupaee2018wikiHowAL}. For each task, we generate an ordered list of steps by extracting the step headlines.

\myparagraph{CrossTask (Evaluation)}
We use this established instructional video benchmark for \emph{zero-shot} grounding, i.e., by directly evaluating on CrossTask our model learned from HowTo100M.
Following common practices, we use two evaluation protocols:
the first one -- \emph{step localization} -- aims at predicting a single timestamp for each occurring step in videos from $18$ primary tasks ~\cite{Zhukov_CVPR19}. 
Performance is evaluated by computing the recall (denoted as Avg. R@1) of the most confident prediction for each task and averaging the results over all query steps in a video, where R@1 measures whether the predicted timestamp for a step falls within the ground truth boundaries. We report average results over 20 random sets of 1850 videos~\cite{Zhukov_CVPR19}.
The second task -- \emph{article grounding} -- requires predicting temporal segments for each step of an instructional article describing the task represented in the video. We use the mapping between CrossTask and \emph{simplified} wikiHow article steps provided in Chen et al.~\cite{Chen_EMNLP22} and report
results on 2407 videos of 15 primary tasks obtained
excluding three primary tasks following the protocol of~\cite{Chen_EMNLP22} (see \supp for details). Performance for this task is measured with Recall@$K$ at different IoU thresholds~\cite{Chen_EMNLP22}. 

\myparagraph{HT-Step (Evaluation)} To evaluate the effectiveness of our model in grounding steps,
we introduce an evaluation benchmark consisting of $1200$ HowTo100M videos spanning a total of $177$ unique tasks, with each video manually annotated with temporal segments for each occurring step. 
For each video, annotators were provided with the task name (e.g., Make Pumpkin Puree) and the recipe steps from the corresponding \href{https://www.wikiHow.life/Make-Pumpkin-Puree}{wikiHow article}.
We refer the reader to \supp for details about the data annotation.
We split the annotated videos into a validation and a test set, each containing 600 videos, with 5 videos per task. We ensure that our validation set does not contain videos from HTM-Align.

\myparagraph{HTM-Align (Evaluation)} This benchmark is used to evaluate our model on narration grounding. It contains 80 videos where the ASR transcriptions have been manually aligned temporally with the video. We report the R@1 metric~\cite{Han_CVPR22}, which evaluates whether the model can correctly localize the narrations that are alignable with the video.

\subsection{Implementation Details} \label{sec:implementation_details}
As  video encoder we adopt the
S3D~\cite{xie2018rethinking} backbone pretrained with the MIL-NCE objective on HowTo100M~\cite{miech20endtoend}.
Following previous work~\cite{xu-etal-2021-videoclip,Han_CVPR22}, we keep this module frozen and use it to extract clip-level features (one feature per second for video decoded at 16 fps).
For extracting context-aware features for each sentence (step or narration), we follow the
Bag-of-word (BoW) approach based on Word2Vec embeddings~\cite{mikolov13}. Our methods hyperparameters were selected on the HT-Step validation set and are: $\lambda_{SV}=\lambda_{NV}=1$, temperatures $\eta,\xi=0.07$, and threshold $\gamma=0.65$. We train our model for $12$ epochs, with 3 epochs burn-in and then we update the teacher every 3 epochs. Pseudo-labels are obtained based on the steps-to-video alignment matrix. To obtain temporal segment detections from the step-to-video alignment output of our model (e.g. for evaluating on the CrossTask article grounding setting) we use a simple 1D blob detector~\cite{wang23egoonly}. Unless otherwise specified, we use the fused alignment matrix for step grounding when narrations are available during inference time.
More details are included in \supp.

\begin{table}[ht]
\centering
\scriptsize
\setlength{\tabcolsep}{1pt} %
\begin{tabular}{lccccc}
\toprule
\multirow{2}{*}{Method} & \multirow{2}{*}{Train. Inp.} & \multicolumn{2}{c}{\textbf{HT-Step $\uparrow$R@1}} & \multicolumn{1}{c}{\textbf{HTM-Align $\uparrow$R@1} } \\
\cmidrule(lr){3-4}
 &  &  w/o nar. & w/ nar. &   \\
 \midrule
CLIP (ViT-B/32)~\cite{radford2021learning} & -  & - & - & 23.4 \\
MIL-NCE~\cite{miech20endtoend}    & N & {\underline{30.7}}  & -  & 34.2 \\
TAN (Joint+Dual, S2)~\cite{Han_CVPR22} & N & -  & - & \underline{49.4} \\
TAN* (Joint, S1, LC)~\cite{Han_CVPR22} & N & 31.2  & - & 47.1 \\
TAN* (Joint, S1, PE+LC)~\cite{Han_CVPR22} & N & 7.9  & - & 63.0 \\
\midrule
Ours  & N+S & $\textbf{35.6} \pm 0.4$ & $\textbf{37.4} \pm 0.4$ & $\textbf{66.5} \pm 0.9$  \\
\bottomrule
\end{tabular}
\caption{\textbf{Comparison with state-of-the-art methods for step and narration grounding.} We report results on the HT-Step and HTM-Align test sets, respectively. TAN* refers to our improved baselines of ~\cite{Han_CVPR22}. 
S1 and S2 refer to the training stages followed in~\cite{Han_CVPR22}. PE denotes the addition of positional encoding to the output of the narration encoder. LC denotes long context, \ie,~our improved TAN* baseline using $1024$ seconds of context as opposed to $64$ for TAN. 
Previous best results are shown underlined. Our \ac{MSTAN} results are reported after 5 random runs. VINA clearly outperforms all previous work -- as well as our improved TAN baselines --  by large margins on both narration alignment and step grounding. }
\label{tab:htm_sota}
\end{table}
\subsection{Results}
\subsubsection{Comparison with the State of the Art}
\vspace{-0.3cm}
\myparagraph{Weakly-Supervised Narration and Step Grounding} Table~\ref{tab:htm_sota} compares the step and narration grounding performance of our method with recent state-of-the-art video-text alignment methods trained on HowTo100M using ASR narrations: MIL-NCE~\cite{miech20endtoend} and TAN~\cite{Han_CVPR22}. When using them for narration alignment, we feed them with ASR as input. But we also evaluate them as strong baselines for zero-shot step grounding by feeding them with the sequence of steps as input. Our model achieves $66.5\%$ R@1 on narration alignment on HTM-Align, leading
to an absolute improvement of $17.1\%$
over the previously reported state-of-the-art (49.4\%). 
Notably, on HTM-Align our method surpasses \emph{TAN* (Joint, S1, PE, LC)} which is a new version of TAN~\cite{Han_CVPR22} implemented by us and much stronger in video-narration alignment. Our re-implementation uses positional encodings for ASR narrations, is trained on long-form videos (up to 17 minutes) only with original ASR timestamps, while TAN was trained on 1 min video-clips with refined narration timestamps and used a fusion of two models during inference (Joint+Dual). 
Our method also outperforms all baselines for step grounding on HT-Step even when seeing only steps during inference, while being trained with (video, narrations, steps) triplets. It also outperforms TAN* (Joint, S1, LC), which is a second re-implementation of TAN designed for maximum performance on the task of step grounding.
Additionally, \ac{MSTAN} is able to use ASR transcripts of videos during test time, if available, to further boost the performance.

\paragraph{Step localization on CrossTask.} 

\begin{table}[ht]
\centering
\footnotesize
\setlength{\tabcolsep}{6pt} %
\label{tab:sota-comparison}
\begin{tabular}{lcccc}
\toprule
Method & $\uparrow$Avg. R@1 (\%) \\
\midrule
\multicolumn{2}{l}{\emph{Supervised}} \\
TempCLR~\cite{yang2021temporal}   & 52.5 \\
\midrule
\multicolumn{2}{l}{\emph{Zero-Shot}} \\
HT100M~\cite{miech2019howto100m}   & 33.6  \\
VideoCLIP~\cite{xu-etal-2021-videoclip}   & 33.9 \\
MCN~\cite{chen2021multimodal}   & 35.1 \\
DWSA~\cite{shen2021learning}   & 35.3 \\
MIL-NCE~\cite{miech20endtoend} &  40.5 \\
Zhukov~\cite{Zhukov_CVPR19} &  40.5 \\
VT-TWINS*~\cite{ko2022video}   &  40.7 \\
UniVL~\cite{Luo2020UniVL}   & 42.0 \\
\midrule
Ours w/o nar.   & \textbf{44.1} \\
Ours w/ nar.    & \textbf{44.8} \\
\bottomrule
\end{tabular}
\caption{\textbf{Comparison with state-of-the-art methods for zero-shot action step localization on the CrossTask dataset}.
The performance of the state-of-the-art fully-supervised method (TempCLR~\cite{yang2021temporal}) is reported as an upper-bound to the zero-shot approaches. * denotes results reported on different test splits, and hence not directly comparable with the rest.
Our model outperforms all previous works by a clear margin (2.1\% absolute improvement over the previous best result on the standard  split). When providing narrations as additional inputs during inference (only text, not the timings), we obtain a further $0.7\%$ boost.  
}
\label{tab:crosstask_step_localization}
\end{table}

In \tbl{tab:crosstask_step_localization} we compare our model against the state-of-the art in step localization on the CrossTask benchmark. Our approach sets a new state-of-the-art for zero-shot step localization on this challenging benchmark. Importantly, most approaches are evaluated on this dataset by feeding their predicted steps-to-frames alignment matrix to a dynamic programming algorithm which finds the optimal assignment of each step with exactly one short clip \emph{assuming a canonical, fixed ordering of steps for each task}.
In contrast, our method, which is naturally aware of context and ordering by densely grounding steps, can outperform prior results without imposing any constraints during inference.

\begin{table}[h]
\setlength{\tabcolsep}{6pt} %
\centering
\footnotesize
\label{tab:results}
\begin{tabular}{lccc|ccc}
\toprule
\multirow{2}{*}{Model} & \multicolumn{3}{c|}{\textbf{$\uparrow$R@50(IOU)}} & \multicolumn{3}{c}{\textbf{$\uparrow$R@100(IOU)}} \\
\cmidrule(lr){2-4} \cmidrule(lr){5-7}
& 0.1 & 0.3 & 0.5 & 0.1 & 0.3 & 0.5 \\
\midrule
 MIL-NCE-max~\cite{miech20endtoend} &   33.5    &  12.0  & 4.9 & 39.7 & 14.3 & 5.9 \\
 MIL-NCE-avg~\cite{miech20endtoend} & 42.9  &  24.3  & 12.9 & 56.8  & 32.1  & 17.0  \\
 WSAG~\cite{Chen_EMNLP22}       & 40.1  &  23.1  & 10.1 & 54.3  & 31.3  & 14.0  \\
\midrule
Ours                            &  \textbf{87.1}   &  \textbf{59.0}  & \textbf{30.0} & \textbf{90.6} & \textbf{61.1} & \textbf{30.9} \\
\midrule
\end{tabular}
\caption{\textbf{Comparison with state-of-the-art approaches for article grounding on the CrossTask dataset.}
}
\label{tab:crosstask_article_grounding}
\vspace{-20pt}
\end{table}

\paragraph{Article grounding on CrossTask.} \ac{MSTAN} is robust to the type of language in which task steps are described. It can handle both atomic phrases (as demonstrated by our results on step localization on CrossTask), but also rich, natural language step descriptions, as evidenced by performance on HT-Step. To further demonstrate this, we compare against the state-of-the art on the article grounding task of CrossTask in \tbl{tab:crosstask_article_grounding}. Our model outperforms all previous works by a large margin. We emphasize the performance improvement we obtain compared to WSAG, which  highlights the importance of exploiting the narration information for training.

\vspace{-10pt}
\subsubsection{Ablation Studies} 
We perform ablations to assess the impact of the various design choices in our method by measuring step grounding performance on the HT-Step validation set and video-narration alignment performance on HTM-Align.

\begin{table}[ht]
\centering
\footnotesize
\setlength{\tabcolsep}{2pt} %
\label{tab:multitask_abl}
\begin{tabular}{clccccc}
\toprule
& \multirow{2}{*}{Method} & \multirow{2}{*}{Train. Inp.} & \multirow{2}{*}{Iter. Pseudo.} & \multicolumn{2}{c}{\textbf{HT-Step $\uparrow$R@1}} & \multicolumn{1}{c}{\textbf{HTM-Align}} \\
\cmidrule(lr){5-6}
 & &  & & w/o nar. & w/ nar. &   \\
\midrule
\multicolumn{7}{c}{\emph{Baseline/Initial Step Pseudo-labels}} \\
(1) & TAN Joint S1  & N &  & 30.7 & - & 47.1 \\
\midrule
\multicolumn{7}{c}{\emph{Single-Task Training}} \\
(2) & Ours  & N & &  - & -  & 63.2 \\
(3) & Ours  & S & & 34.0 & - & - \\
(4) & Ours   & S & \checkmark & 35.8 & - & - \\
\midrule
\multicolumn{7}{c}{\emph{Multi-Task Training}} \\
(5) & Ours & N+S & & 34.3 & 36.1 & 64.8 \\
(6) & Ours & N+S & \checkmark & \textbf{36.9} & \textbf{39.1} & \textbf{67.0} \\
\bottomrule
\end{tabular}
\caption{\textbf{Ablation of main components of our framework}. We study the contribution of (a) multi-task training for narration and step grounding, (b) iterative step pseudo-labeling (\emph{Iter. Pseudo}), and (c) narration-aware step grounding (\emph{w/ nar.}). 
We report results on the HT-Step val set for STG and HTM-Align for NG. 
We compare training only with narrations (N), only with wikiHow steps (S), and training with narrations-steps sequence pairs (N+S). We also compare the performance with and without providing narrations during inference.}
\label{tab:ablation_multimodal}
\vspace{-20pt}
\end{table}

\paragraph{Effect of weak supervision from instructional articles.} Row 3 in Table~\ref{tab:ablation_multimodal} shows the step grounding results obtained from an instance of our model that includes only the direct video-step alignment pathway and that is trained just on wikiHow steps (without narrations) using the fixed step pseudo-labels from TAN*~\cite{Han_CVPR22} without any form of iterative pseudo-labeling (row 1). Remarkably, this variant improves by $3.3\%$ over the step-grounding performance of TAN*. When we let this variant update the step pseudo-labels (row 4), the recall improves further ($5.1\%$ over TAN*). These results provide evidence of the strong benefits of utilizing instructional articles for the learning of step grounding. 

\paragraph{Effect of multimodal training and inference.} Training our model with multi-modal textual inputs (steps and narrations), we observe an improvement of $1.6\%$ in narration grounding (row 5 of Table~\ref{tab:ablation_multimodal}) compared to its single-task counterpart (row 2). However the gain in step grounding is marginal when seeing only video frames during inference (\emph{w/o nar.}, 34\% in row 3 vs 34.3\% in row 5). 
Our conjecture is that the missing modality (narrations) leads to some drop in performance.  Providing both steps and narrations during inference leads to a stronger step grounding performance, which surpasses the TAN* baseline by $5.4\%$ (30.7 $\rightarrow$ 36.1).

\vspace{-10pt}
\paragraph{Effect of iterative pseudo-labeling.}
By comparing row 5 to row 6 of Table~\ref{tab:ablation_multimodal} we observe a clear boost in performance on both step grounding and narration alignment. This is a clear indication of the gains produced by iteratively refining the pseudo-labels using our model as a teacher during training. 

\begin{table}[ht]
\centering
\footnotesize
\begin{tabular}{lccccc}
\toprule
 Alignment & S $\rightarrow$ V & S $\rightarrow$ N & N $\rightarrow$ V & \textbf{HT-Step $\uparrow$R@1} 
 \\
 \midrule
S $\rightarrow$ V & learned & - & - & 34.3 \\
S $\rightarrow$ N $\rightarrow$ V & - & learned & learned & 30.5 \\
S $\rightarrow$ N $\rightarrow$ V & - & learned & ASR  & 27.9 \\
S $\rightarrow$ N $\rightarrow$ V & - & MPNet~\cite{song2020mpnet} & ASR  & 19.0 \\
Fused & learned & learned & learned & \textbf{36.1} \\
\bottomrule
\end{tabular}
\caption{\textbf{Impact of the alignment matrix used during inference with narrations}. The same model is used for all results (corresponding to row 5 in Table~\ref{tab:ablation_multimodal}).}
\vspace{-20pt}
\label{tab:snv_abl}
\end{table}

\paragraph{Impact of pathways during inference.} In \tbl{tab:snv_abl} we study the effects of using different pathways and alignment information during inference. All results are produced from the same model trained for joint narration and step grounding with fixed pseudo-labels from TAN (row 5 in Table~\ref{tab:ablation_multimodal}). Grounding steps using the indirect steps-to-video alignment only lags by $3.8\%$ behind the direct steps-to-video alignment that directly computes the similarity between steps and video frames ($30.5\%$ vs $34.3\%$). Their fusion outperforms their individual grounding performance. This suggests that they capture complementary information. We also explore substituting our learned steps-to-narrations alignment with an alignment computed with an off-the-shelf language model. This significantly degrades performance ($19.0\%$) showing that our joint steps and narrations grounding model learns relationships between steps and narrations that go beyond textual similarity between pairs of sentences. Similarly, substituting our learned narrations-to-video alignment with an alignment based on the original ASR timestamps reduces performance by $2.6\%$.

\paragraph{Iterative pseudo-labeling strategies.}
\vspace{-10pt}
In \tbl{tab:ablation_iterative_pseudolling} we ablate design choices for the iterative pseudo-labeling stage. We can observe that using aggressive filtering (i.e., high thresholds translating to a high maximum percentage of pseudo-labels that are discarded) is key to observing gains from iterative pseudo-labeling (using either the S $\rightarrow$ V or Fused alignment matrices) compared to training with fixed pseudo-labels from TAN. Intuitively, a large percentage of steps described in wikiHow articles are not represented in the given instructional video due to task mismatch, variations in recipe execution, and some steps being optional. Therefore, starting with a small subset of reliable pseudo-labels can facilitate step grounding.

\begin{table}[ht]
\setlength{\tabcolsep}{2pt} %
\footnotesize
\centering
\label{tab:align_filt}
\begin{tabular}{lcccc}
\toprule
\multirow{2}{*}{Alignment} & \multirow{2}{*}{$\gamma$} &  \multirow{2}{*}{max \% step} & \multicolumn{2}{c}{\textbf{HT-Step $\uparrow$R@1}} \\
\cmidrule(lr){4-5}
 &  & discarded &  w/o nar. & w/ nar.   \\
\midrule
S $\rightarrow$ V  & 0.40 & 24 & 34.3 & 36.5  \\
S $\rightarrow$ V  & 0.65 & 91 & \textbf{36.9} & \textbf{39.1}  \\
\midrule
Fused & 0.40 & 60 & 34.1 & 35.9   \\
Fused & 0.55 & 88 &  36.2 & 35.5  \\
\bottomrule
\end{tabular}
\caption{\textbf{Ablation of the type of alignment matrix and filtering threshold used for pseudo-label generation}. Pseudo-label generation with the steps-to-video alignment matrix and the fusion of the direct and indirect pathways perform comparably for step grounding. Aggressive unreliable pseudo-label filtering with high confidence thresholds $\gamma$ (large maximum step discard ratio) helps in both cases.
}
\label{tab:ablation_iterative_pseudolling}
\end{table}

\paragraph{Task selection.} 
\vspace{-10pt}

\begin{table}[ht]
\centering
\footnotesize
\begin{tabular}{lc}
\toprule
Task ID selection & \textbf{HT-Step $\uparrow$R@1} \\
\midrule
HT100M metadata      & 34.3   \\ 
Top-1 prediction     & 34.7   \\ 
Random / top-5 pred  & 34.3   \\ 
\bottomrule
\end{tabular}
\caption{ \textbf{Sensitivity to task id selection.} 
We assess how the performance of our method changes when using different
strategies to associate videos with articles. We experiment with using the task ids
available from HT100M, as well as the the two predictive strategies presented
in Section~\ref{subsec:method_task_inf}. We conclude that our method is robust to the task selection, and the task labels are not necessary for training.  
}
\label{tab:ablation_taskid}
\vspace{-10pt}
\end{table}

In \tbl{tab:ablation_taskid} we investigate different strategies to select the wikiHow articles during training. This selection determines the set of steps to be grounded. We evaluate two strategies for video-task association from narrations and compare them with using the task id provided in the HowTo100M metadata for each video\footnote{During inference, metadata task ids are used in all of our HowTo100M experiments in order to evaluate against the ground-truth step annotations.}. We see that our automatic selection approaches yield results on par with or even slightly better than those based on metadata. 

\vspace{-5pt}
\section{Conclusion}
We have presented a method for learning how to temporally
ground sequences of steps in instructional videos, without any manual supervision. Our proposed method exploits the weak supervision naturally provided
in such videos through their narrations,
and solves for joint alignment of narrations and steps, while fusing two complementary pathways for
step-to-video alignment. We demonstrated strong quantitative
performance, surpassing the state-of-the-art on multiple benchmarks for both narration and step grounding.

\myparagraph{Acknowledgements}
We thank Huiyu Wang, Yale Song, Mandy Toh, and Tengda Han for helpful discussions.

\newpage
\appendix

This Appendix provides: additional details (annotation procedure, statistics) about the HT-Step dataset that we introduced for evaluating models on step grounding (Section~\ref{sup:htm_step}), additional details for the rest of the datasets that were used for training/evaluation (Section~\ref{sup:other_datasets}), implementation details (Section~\ref{sup:impl}), qualitative results for step grounding on HT-Step (Section ~\ref{sup:qual}), additional ablation studies (Section ~\ref{sup:ablations}), and additional details about the evaluation of our models on HTM-Align (Section ~\ref{sup:htm_align}).

\section{HT-Step Dataset}
\label{sup:htm_step}
\begin{table}[h]
\centering
\footnotesize
\setlength{\tabcolsep}{2pt} %
\begin{tabular}{lccccc}
\toprule
Dataset & Step Annot. & \# Videos & \# Activities & \# Steps &  \# Segments \\
\midrule
HowTo100M~\cite{miech2019howto100m} & \ding{55} & 1.2M & 25k & - & - \\
HTM-Align~\cite{Han_CVPR22} & \ding{55} & 80 & 80 & - & - \\
CrossTask~\cite{Zhukov_CVPR19} & \checkmark & 4.8k (2.8k) & 83 (18) & 133 & 20.9k \\
HT-Step (val) & \checkmark & 600 & 120 & 1,204 & 3,441 \\
HT-Step (test) & \checkmark & 600 & 120 & 1,242 & 3,631 \\
\midrule
wikiHow &  & - & 14k & 100k & - \\
\bottomrule
\end{tabular}
\caption{Summary statistics for the datasets in our work. For CrossTask, the statistics for primary activities only are shown in parentheses.}
\label{tab:dataset_stats}
\end{table}

In this section we provide details about the creation of the HT-Step benchmark that we used for evaluating our models.
This benchmark was designed to provide a high-quality set of 
step-annotated instructional videos for
a plethora of tasks, described in rich, structured language instead of atomic phrases.

\myparagraph{Annotation setup} 
We used videos from the HowTo100M dataset; each one of those videos contains a task id label that corresponds to a wikiHow article. 
This association enabled us to obtain a set of potential step descriptions for every video, directly from the corresponding wikiHow article. 
We note that this association is noisy, e.g. the video might show a variation of a specific recipe, where some of the steps in the article often do not appear at all, appear partially, are executed in different order, or are repeated multiple times.

\myparagraph{Annotation instructions} 
For each video, annotators were provided with the task name (e.g., Make Pumpkin Puree) and the recipe steps from the corresponding \href{https://www.wikiHow.life/Make-Pumpkin-Puree}{wikiHow article}.
The annotators where asked to watch the whole video and first decide whether it is relevant to the given task -- i.e. if at least some of the given steps were visually demonstrated and the task's end goal was the same (e.g. a specific recipe) -- or reject it otherwise. 
When a video was deemed relevant, annotators were asked to mark all instances of the provided steps with a temporal window.
We note that WikiHow articles often contain several variations/methods for completing a given task. For tasks where this was the case, the annotators were asked to select the set of steps corresponding to the variation that best fits every video and only use those steps for annotating the entire video. 

\myparagraph{QA process} 
To ensure the quality of the annotations, we followed a rigorous multi-stage Quality Assurance (QA) process:
In the first stage, the videos were annotated by a single annotator.
These initial annotations were then reviewed by more experienced annotators, who either approved all the annotations on a given video (meaning all the marked steps were correct and no steps were missing) or marked it for redoing with specific comments on which annotations needed fixing and in what way. 
At the last stage of the QA process, the annotations that were marked as incorrect were redone by third, independent annotators.

\myparagraph{Statistics}
We provide per-activity statistics for the annotations in \tbl{tbl:htm_step_stats}. 
The metrics used, \ie~number of unique steps, step and video coverage,
are given to provide an understanding of how the number of steps varies between different tasks and how the steps of a task may appear partially 
in the HowTo100M videos.   

\myparagraph{Validation and test (val/test) split} 
Overall during the full annotation process, approximately $35\%$ of the videos were rejected as irrelevant to the given tasks. 
We split the remaning, annotated videos into a validation and a test set, each containing 600 videos, with 5 videos per task. We ensured that our validation set does not contain videos from HTM-Align.
In total 87 human annotators manually annotated $1200$ videos over $177$ tasks: $120$ in the validation and $120$ in the test set, with $5$ videos per task, i.e. with $63$ tasks overlapping between the two sets.

\begin{table}[h]
\centering
\scriptsize
\setlength{\tabcolsep}{2pt} %
\begin{tabular}{lcccc}
\toprule
\textbf{Task} & \textbf{\# steps} & \textbf{step coverage} & \textbf{video coverage}  \\
\midrule
Make Zucchini Pancakes&4.0&0.83&0.37                       \\
Make a Hearty Stew&3.5&0.82&0.12                           \\
Make Beef and Broccoli&3.1&0.78&0.24                       \\
Make Coconut Popsicles&3.8&0.76&0.28                       \\
Make Yorkshire Pudding&5.3&0.76&0.11                       \\
Cook Spaghetti alla Carbonara&4.6&0.73&0.39                \\
Make Vegan Pesto&2.2&0.73&0.15                             \\
Make Corn Fritters&6.4&0.72&0.28                           \\
Make Buttermilk Fried Chicken&4.2&0.70&0.44                \\
Make a Shrimp Po Boy Sandwich&4.2&0.70&0.27                \\
$\vdots$ & $\vdots$ & $\vdots$ & $\vdots$ & $\vdots$ \\
Cook Prime Rib&2.6&0.19&0.19                   \\
Cure Bacon&2.2&0.18&0.11                       \\
Make Dim Sum&4.6&0.18&0.15                     \\
Make Vegan Ceviche&2.8&0.17&0.08               \\
Make Lobster Bisque&3.6&0.17&0.28              \\
Make Giblet Gravy&2.8&0.16&0.23                \\
Make Pickled Eggs&4.4&0.16&0.19                \\
Pickle Onions&1.6&0.15&0.12                    \\
Cook Rib Eye Roast&2.0&0.12&0.28               \\
Make Pap&2.0&0.12&0.21                         \\
\midrule
\textbf{Average}           & \textbf{4.0}	& \textbf{0.42}	& \textbf{0.24} \\
\bottomrule
\end{tabular}
\caption{
Statistics of the annotations used to create the HT-Step benchmark. The metrics are computed per task (for $177$ tasks in total), averaged over all the annotated videos for a given task. \textbf{\# steps} denotes the average number of unique steps annotated per video, per activity; \textbf{step coverage} denotes the fraction of a task's steps that have been found and annotated in every video; \textbf{video coverage} denotes the fraction of the video's duration that is covered by step annotations; 
Rows are sorted by step coverage; only the 10 tasks with the highest and lowest step coverage are shown here for brevity. 
}
\label{tbl:htm_step_stats}
\end{table}

\section{Datasets Details}
\label{sup:other_datasets}
We provides a statistics summary for the datasets used for training and evaluation in \tbl{tab:dataset_stats}.

\myparagraph{HowTo100M (Training)} 
HowTo100M contains over 1M unique instructional videos, spanning over 24k activities including cooking, DIY, arts and crafts, gardening, personal care, fitness and more. Each instructional video is complemented by the ASR transcription of it's audio, which usually contains the real time narration/commentary of the instructor during the activity. 
We use the "senticified" version of the ASR sentences provided by Han~\etal~\cite{Han_CVPR22}.
Following Han~\etal~\cite{Han_CVPR22} we also train only using the Food \& Enteratainment subset, which includes a subset of approximately $370k$ videos. 

\myparagraph{wikiHow (Training)} We train using 14,541 cooking tasks from the wikiHow-Dataset~\cite{Koupaee2018wikiHowAL}. For each task, we generate an ordered list of steps by extracting the step headlines.
The HowTo100M dataset was curated using a semi-automatic pipeline that involved searching YouTube with queries based on the titles of wikiHow articles. 
Consequently there is an almost complete overlap in activities between the two corpora, which makes wikiHow a natural choice for mining step-level articles to associate with instructions in HowTo100M videos.  
In the context of this paper we used the wikiHow-Dataset~\cite{Koupaee2018wikiHowAL} to collect the articles for 14,541 cooking tasks.  

\myparagraph{CrossTask (Evaluation)}
We use this established instructional video benchmark for \emph{zero-shot} grounding, i.e., by directly evaluating on CrossTask our model learned from HowTo100M.
The Crosstask dataset~\cite{Zhukov_CVPR19}. is an established benchmark for temporal localization of steps in instructional videos. It consists of 4800 videos from 83 activities, which
are divided into $18$ primary ($14$ related to cooking
and $4$ to DIY car repairs and shelf assembly) and $65$ related activities.
The videos in the primary activities are annotated with step annotations
in the form of temporal segments from a predefined taxonomy of $133$ steps. Those steps tend to be atomic, \eg~ for activity ``Make Taco Salad" the available steps are ``add onion", ``add taco", ``add lettuce", ``add meat", ``add tomato", ``add cheese", ``stir", and ``add tortilla".
Following common practices, we use two evaluation protocols:
the first one -- \emph{step localization} -- aims at predicting a single timestamp for each occurring step in videos from $18$ primary tasks ~\cite{Zhukov_CVPR19}. 
Performance is evaluated by computing the recall (denoted as Avg. R@1) of the most confident prediction for each task and averaging the results over all query steps in a video, where R@1 measures whether the predicted timestamp for a step falls within the ground truth boundaries. We report average results over 20 random sets of 1850 videos~\cite{Zhukov_CVPR19}.
The second task -- \emph{article grounding} -- requires predicting temporal segments for each step of an instructional article describing the task represented in the video. We use the mapping between CrossTask and \emph{simplified} wikiHow article steps provided in Chen et al.~\cite{Chen_EMNLP22} and report
results on 2407 videos of 15 primary tasks obtained
excluding three primary tasks following the protocol of~\cite{Chen_EMNLP22}. Performance for this task is measured with Recall@$K$ at different IoU thresholds~\cite{Chen_EMNLP22}. 

\myparagraph{HTM-Align (Evaluation)} This benchmark is used to evaluate our model on narration grounding. It contains 80 videos where the ASR transcriptions have been manually aligned temporally with the video. In the main submission, we report the R@1 metric~\cite{Han_CVPR22}, which evaluates whether the model can correctly localize the narrations that are alignable with the video. In Section~\ref{sup:htm_align} we also evaluate our model in terms of its capability to decide whether a narration is visually groundable in the video or not using the ROC-AUC metric~\cite{Han_CVPR22}. AUC denotes the area the ROC curve of the alignment task, and measures the ability of the model to correctly predict whether a given step is alignable within a video or not. 

\section{Implementation Details}
\label{sup:impl}
As  video encoder we adopt the
S3D~\cite{xie2018rethinking} backbone pretrained with the MIL-NCE objective on HowTo100M~\cite{miech20endtoend}. Following previous work~\cite{xu-etal-2021-videoclip,Han_CVPR22}, we keep this module frozen and use it to extract clip-level features (one feature per second for video decoded at 16 fps). For extracting context-aware features for each sentence (step or narration), we follow the
Bag-of-word (BoW) approach based on Word2Vec embeddings~\cite{mikolov13}. These embeddings are initialized based on MIL-NCE Word2Vec and are fine-tuned during training.

The hyperparameters of the model compared with state-of-the-art methods in Tables 1,2,3 of the main submission were selected based on R@1 performance on the HT-Step validation set and are: $\lambda_{SV}=\lambda_{NV}=1$, temperatures $\eta,\xi=0.07$, and pseudo-label filtering threshold $\gamma=0.65$. We train our model for $12$ epochs, with 3 epochs burn-in training with step pseudo-labels generated by TAN, and then we update the teacher VINA every 3 epochs. We use the AdamW~\cite{loshchilov2018decoupled} optimizer, having an initial learning rate of $2e-4$ decayed with a cosine learning schedule. Our batch size is $32$ videos, with maximum length of $1024$ seconds. 

Pseudo-labels are obtained based on the steps-to-video alignment matrix and are generated (before filtering) as follows: for each step we find the timestep with maximum similarity with the step and then extend a temporal segment to the left and right of that peak as long as the similarity score does not follow below $0.7$ of the peak height. Pseudo-labels whose peak score falls below the filtering threshold $\gamma$ are not used for training.

The rest of hyperparameters were selected based on TAN~\cite{Han_CVPR22}. The multimodal encoder is a pre-norm multi-layer transformer which consists of $6$ layers of self-attention, with $8$ heads and has hidden dimension $D=512$. A learnable positional encoding of size $D=512$ is used to inject temporal information to each frame/narration/step token.

To obtain temporal segment detections from the step-to-video alignment output of our model (e.g. for evaluating on the CrossTask article grounding setting or for the qualitative video included in this supplementary) we use a simple 1D blob detector~\cite{wang23egoonly}. Unless otherwise specified, we use the fused alignment matrix for step grounding when narrations are available during inference time.

Our model is trained on 8 GPUs (Tesla V100-SXM2-32GB) and training lasts approximately 10-12 hours. All models were implemented in Python using Pytorch and are 
based on the PySlowFast (\url{https://github.com/facebookresearch/SlowFast}) and TAN (\url{https://github.com/TengdaHan/TemporalAlignNet}) open-source codebases. For ablation studies, we choose the best checkpoint for each configuration based on performance on HT-Step validation set and report its test split performance.

\section{Qualitative Results}
\label{sup:qual}

\begin{figure*}[ht]
\begin{subfigure}[b]{1.0\textwidth}
\includegraphics[width=\textwidth]
{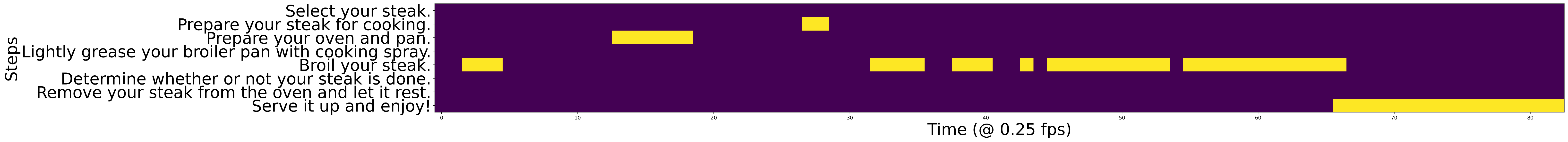}
\caption{Ground-truth step grounding annotations.}
\end{subfigure}
\begin{subfigure}[b]{1.0\textwidth}
\includegraphics[width=\textwidth]
{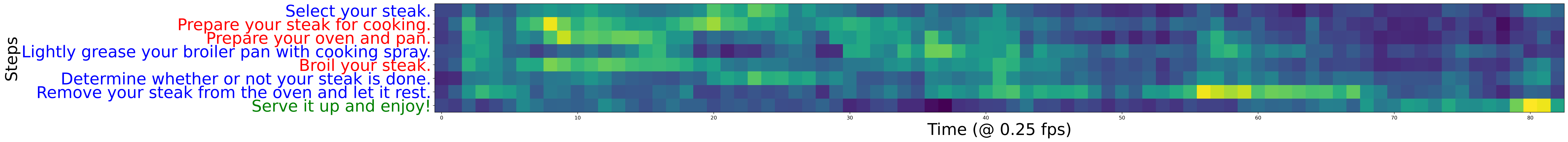}
\caption{Predicted alignment by TAN*.}
\end{subfigure}
\begin{subfigure}[b]{1.0\textwidth}
\includegraphics[width=\textwidth]
{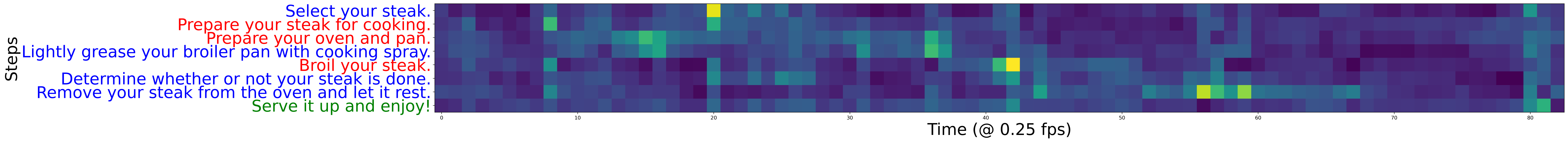}
\caption{Predicted alignment by VINA w/o narrations.}
\end{subfigure}
\begin{subfigure}[b]{1.0\textwidth}
\includegraphics[width=\textwidth]
{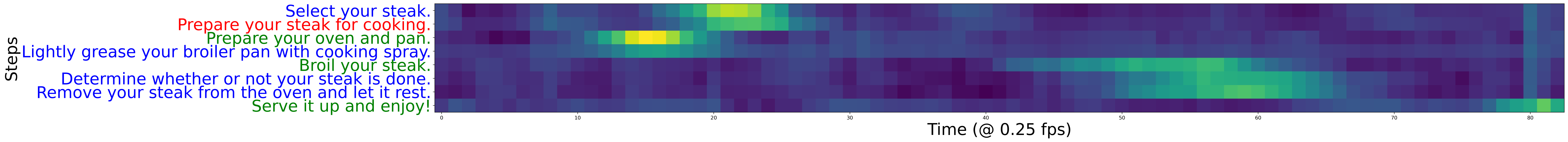}
\caption{VINA w/ narrations.}
\end{subfigure}
\caption{Qualitative results on a validation video of the HT-Step dataset (\texttt{VIQYQkA3mNU}) demonstrating how to \emph{Broil Steak}. Steps that are not visually groundable in the video are highlighted in {\color{blue} blue}, steps that are correctly retrieved by each model are highlighted in {\color{green} green}, while steps that are not retrieved are shown in {\color{red}{red}}. Figure best viewed zoomed in and in color.}
\label{fig:not_in_order}
\end{figure*}

\begin{figure*}[ht]
\begin{subfigure}[b]{1.0\textwidth}
\includegraphics[width=\textwidth]
{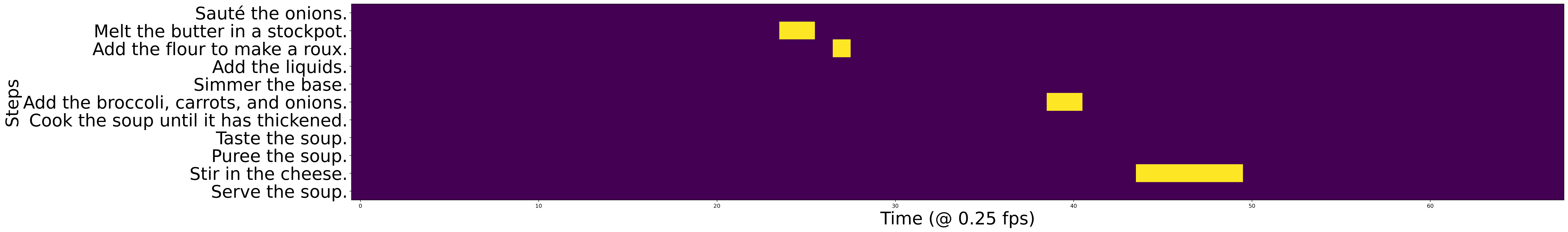}
\caption{Ground-truth step grounding annotations.}
\end{subfigure}
\begin{subfigure}[b]{1.0\textwidth}
\includegraphics[width=\textwidth]
{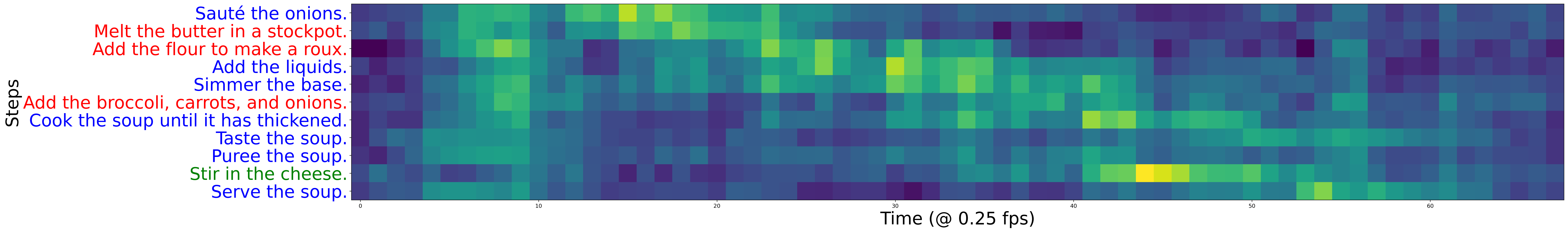}
\caption{Predicted alignment by TAN*.}
\end{subfigure}
\begin{subfigure}[b]{1.0\textwidth}
\includegraphics[width=\textwidth]
{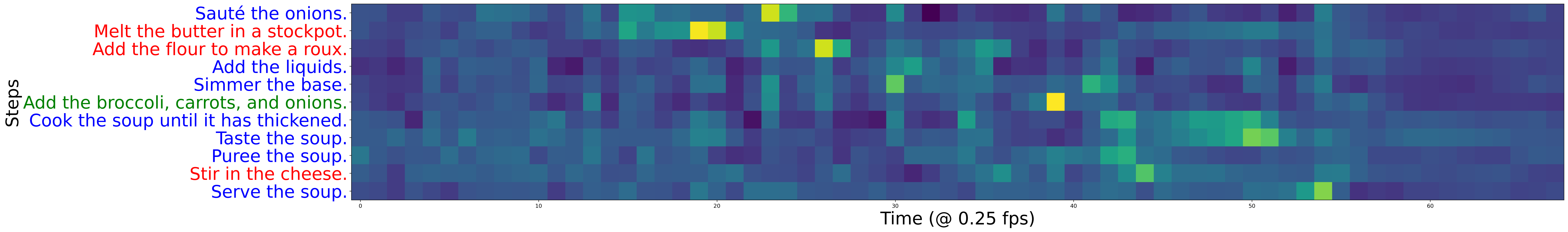}
\caption{Predicted alignment by VINA w/o narrations.}
\end{subfigure}
\begin{subfigure}[b]{1.0\textwidth}
\includegraphics[width=\textwidth]
{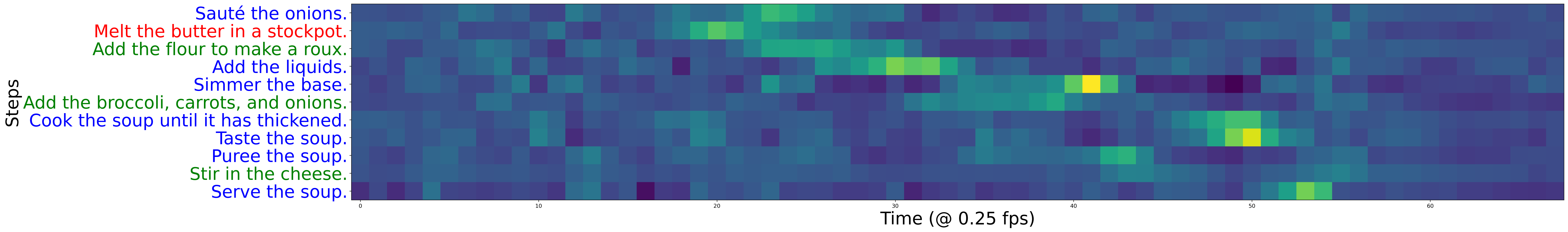}
\caption{VINA w/ narrations.}
\end{subfigure}
\caption{Qualitative results on a validation video of the HTM-Step dataset (\texttt{0dHofx1lqAg}) demonstrating how to \emph{Make Broccoli Cheese Soup}. Steps that are not visually groundable in the video are highlighted in {\color{blue} blue}, steps that are correctly retrieved by each model are highlighted in {\color{green} green}, while steps that are not retrieved are shown in {\color{red}{red}}. Figure best viewed zoomed in and in color.}
\label{fig:nar_help_2}
\end{figure*}

In this section, we provide qualitative results for the ground-truth steps-to-video alignment and predicted alignments by our improved baseline that serves as the initial teacher model (TAN*), and our model (using the direct steps-to-video alignment without narrations) or the fusion with the indirect steps-to-video alignment (with narrations). From these qualitative results (\fig{fig:not_in_order}), we observe that our VINA model can correctly temporally localize visually groundable steps, despite being trained only with noisy pairs of narrated videos and instructional steps. Predicted alignments tend to also be less noisy than TAN*, showcasing the effectiveness of training a video-language alignment model with distant supervision from WikiHow articles. Our model can also leverage ASR transcripts (without any temporal information regarding when the instructor uttered each narration) to further improve its results (\fig{fig:nar_help_2}). 

\section{Extra Ablations}
\label{sup:ablations}

\paragraph{Architecture ablations.} In \tbl{tab:arch_abl} we study the design of the unimodal encoder used to embed steps before they are fed to our Multimodal Transformer. Overall, using positional embeddings capturing the ordering of steps in a task, and using modality-specific projection MLPs leads to a slightly better performance in step grounding (w/o narration input). Narration grounding seems to benefit from using a shared text encoder, possibly because this facilitates knowledge transfer from the WikiHow steps. 

\begin{table}[ht]
\centering
\footnotesize
\begin{tabular}{ccccc}
\toprule
\multirow{2}{*}{PE} & \multirow{2}{*}{Sep. MLP} & \multicolumn{2}{c}{\textbf{HT-Step $\uparrow$R@1}} & \multicolumn{1}{c}{\textbf{HTM-Align}} \\
\cmidrule(lr){3-4}
 &  & w/o nar. & w/ nar. &   \\
\midrule
  & \checkmark & 33.5  & 34.0 & 65.8 \\
 &  & 34.0 & 34.9 & 65.9\\
 \checkmark &  & 33.8 & 34.4 & \textbf{67.0} \\
 \checkmark &  \checkmark & \textbf{34.3} & \textbf{36.1} & 64.8 \\
\bottomrule
\end{tabular}
\caption{\textbf{Ablation study on architecture design}. We study the contribution of positional encodings for steps (\emph{PE}) and of specialized text projection layers for wikiHow article steps (\emph{Sep. MLP}). All models are trained for joint narration and step grounding with fixed pseudo-labels from TAN and evaluated on HT-Step val split (last row corresponds to row 5 in Table 4 of the main text).}
\label{tab:arch_abl}
\end{table}

\section{Experimental Setup on HTM-Align}
\label{sup:htm_align}
As explained in the official code repository of TAN~\cite{Han_CVPR22} (\url{https://github.com/TengdaHan/TemporalAlignNet/tree/main/htm_align}), the results reported for HTM-Align are obtained with a text moving window of 1 minute, i.e., for each 1-minute temporal segment only ASR captions whose original time-stamps fall within a 3-min window centered around this temporal segment are considered for grounding. Instead, for all our reported results  (for TAN* and VINA) we operate in the more challenging setup where an ASR caption can be grounded in any timestep of the original video (there is no knowledge about the original ASR timestamps during inference). Under this more challenging setup, our model outperforms TAN both in narration retrieval, as measured by Recall@1 (66.5\% vs 49.4\%, as seen in Table 1 of the main submission).  

Our model also performs comparably with TAN in step alignability prediction, as measured by ROC-AUC (76\% vs 75.1\%). Note that our model does not have dedicated alignability head for predicting whether a narration exists or not in the video as TAN~\cite{Han_CVPR22}. Instead, we simply obtain an alignability score by using the maximum cosine similarity score over time, where cosine similarities of each narration with each video frame are computed based on the outputs of the unimodal encoders.

\section{Limitations and Ethical Concerns}
From the qualitative results, we observe that due to the losses used during training, which do not explicitly penalize wrong temporal extent (as long as the predicted heatmap has a peak within the target temporal window), grounded temporal segments tend to be short. This is especially prominent when using the direct steps-to-videos alignment that is explicitly supervised (second to last row of the predicted alignment figures). Furthermore, our training objective does not utilize negative examples, e.g. steps that are not visually groundable, to suppress detections. This can lead to confident detections for missing steps. 
Another limitation of our approach (similar to previous approaches that operate on pre-extracted visual features) is that our performance is limited by the quality of the extracted visual representations. 
Regarding ethical concerns, public instructional video datasets and public knowledge base datasets may have gender, age, geographical and cultural bias.
\end{document}